\documentclass{ieeeaccess}
\usepackage{cite}
\usepackage{amsmath,amssymb,amsfonts}
\usepackage{algorithmic}

\usepackage{graphicx}

\ifCLASSOPTIONcompsoc
\usepackage[caption=true, font=normalsize, labelfont=sf, textfont=sf]{subfig}
\else
\usepackage[caption=true, font=footnotesize]{subfig}
\fi

\usepackage{caption}
\usepackage{textcomp}
\usepackage[table ]{ xcolor}
\usepackage{enumerate}

\def\BibTeX{{\rm B\kern-.05em{\sc i\kern-.025em b}\kern-.08em
    T\kern-.1667em\lower.7ex\hbox{E}\kern-.125emX}}

\begin{document}
\history{Date of publication xxxx 00, 0000, date of current version xxxx 00, 0000.}
\doi{10.1109/ACCESS.2017.DOI}

\title{Facial Makeup Transfer Combining Illumination Transfer}
\author{
	\uppercase{Xin Jin\authorrefmark{1,2,3}},
    \uppercase{Rui Han\authorrefmark{1}}, 
    \uppercase{Ning Ning\authorrefmark{1}}, 
    \uppercase{Xiaodong Li\authorrefmark{1*},and Xiaokun Zhang\authorrefmark{1}}
}

\address[1]{Department of Cyber Security, Beijing Electronic Science and Technology Institute, Beijing, 100070, PR China}
\address[2]{State Key Laboratory of Virtual Reality Technology and Systems, Beihang University, Beijing, 100191, PR China}
\address[3]{Department of Automation, Tsinghua University, Beijing, 100083, PR China}

\markboth
{X. Jin \headeretal: Preparation of Papers for IEEE ACCESS}
{X. Jin \headeretal: Preparation of Papers for IEEE ACCESS}

\corresp{$^*$Corresponding author: Xiaodong Li (e-mail: lxdbesti@163.com).}

\begin{abstract}
To meet the women appearance needs, we present a novel virtual experience approach of facial makeup transfer, developed into windows platform application software. The makeup effects could present on the user’s input image in real time, with an only single reference image. The input image and reference image are divided into three layers by facial feature points landmarked: facial structure layer, facial color layer, and facial detail layer. Except for the above layers are processed by different algorithms to generate output image, we also add illumination transfer, so that the illumination effect of the reference image is automatically transferred to the input image. Our approach has the following three advantages: (1) Black or dark and white facial makeup could be effectively transferred by introducing illumination transfer; (2) Efficiently transfer facial makeup within seconds compared to those methods based on deep learning frameworks; (3) Reference images with the air-bangs could transfer makeup perfectly.
\end{abstract}


\titlepgskip=-15pt

\maketitle

\section{Introduction}
\PARstart{F}{acial} makeup transfer is a new application requirement of virtual reality technology in the image. How to see the virtual makeup effect on the image is the need of many young women. Facial makeup is a technique that changes the appearance with special toiletries such as compact, setting powder, and moisturizer. Under many circumstances, particularly for females, makeup is deemed as a necessary practice to beautify appearance. Emulsions are often used to alter the facial skin detail. Compacts are primarily used to hide defects and overlay the initial facial skin detail. Setting powder often satisfies detail for the skin. Except that, other colour makeup, such as eyeliner and shadow, is applied to the upper layer of the setting powder.

The ever-developing makeup technology now extends to different women facial types, different scenes, different ages, different skin, and even different costumes with different makeup \cite{ref15Chen2011,ref16Chen2013,ref17Jin2010,ref18Chen2012,ref19Chen2015,ref20Chen2011,ref21Jin2016,ref22Chen2014,ref23Jin2018,ref24Jin2017,ref25Zhou2019,ref26Zhou2018,ref27Zhou2018,ref28Zhou2016,ref29Lu2014,ref30Lu2018,ref31Lu2018,ref32Lu2019,ref33Lu2018}. The choice of makeup naturally creates a personal experience but greatly consumes time and damages women's skin. 

\begin{figure*}[htbp]	
	\centering
	\subfloat[Input Image]{\label{fig:Fig1-1}\includegraphics[width=0.18\linewidth]{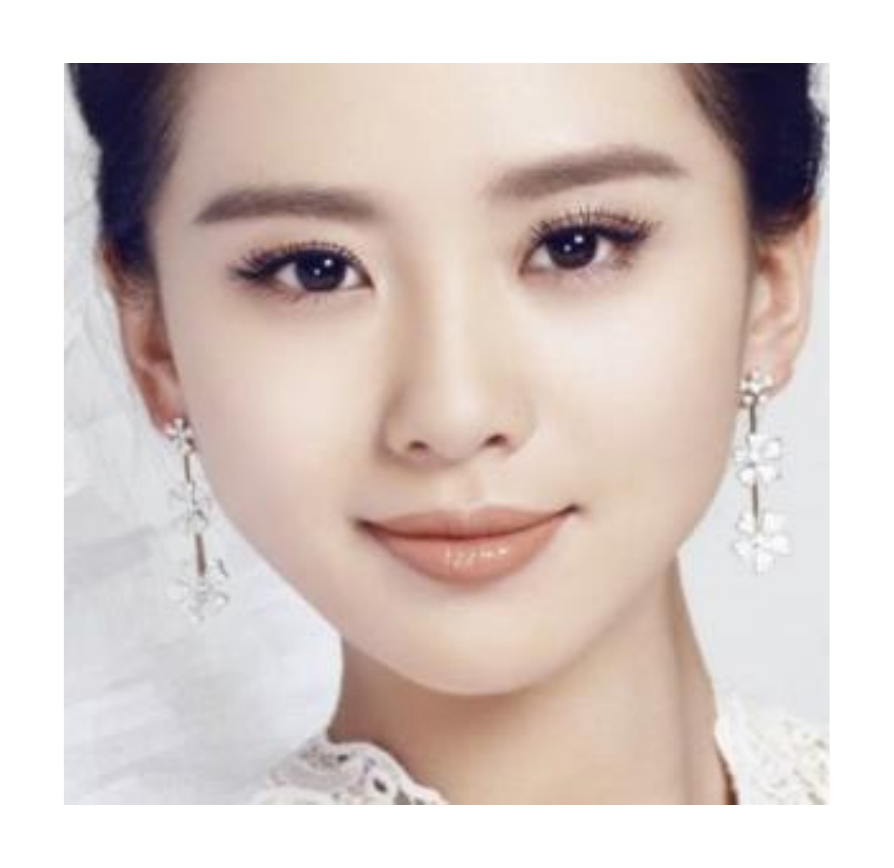}}
	\subfloat[Reference Image]{\label{fig:Fig1-2}\includegraphics[width=0.18\linewidth]{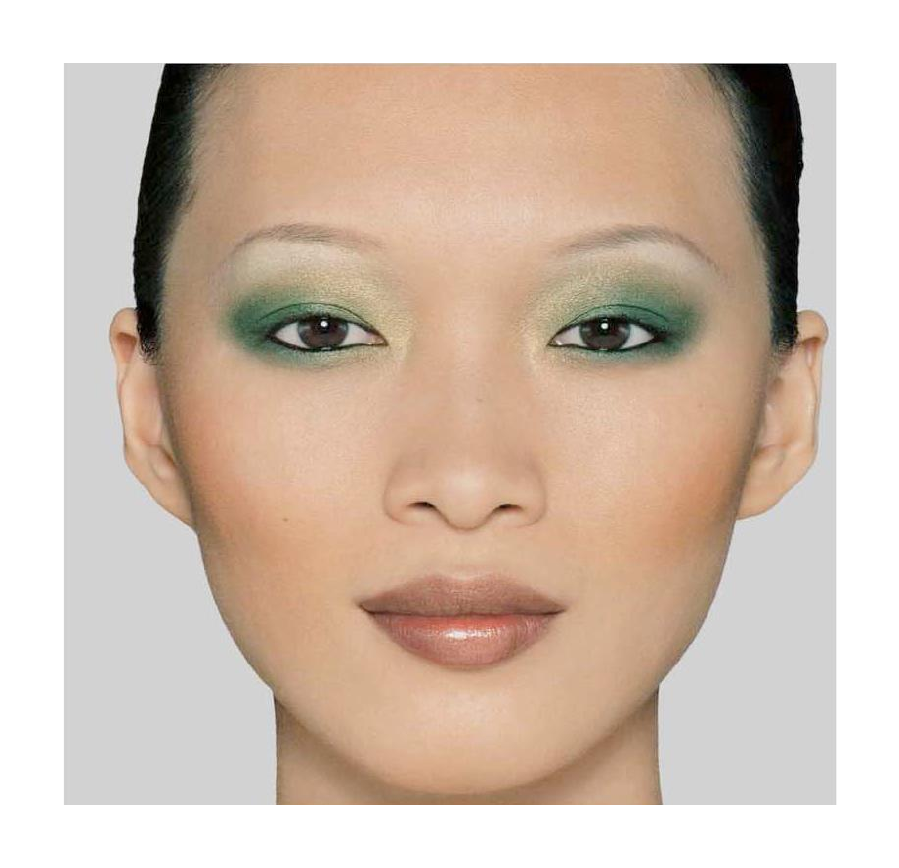}}
	\subfloat[Output Image ]{\label{fig:Fig1-3}\includegraphics[width=0.18\linewidth]{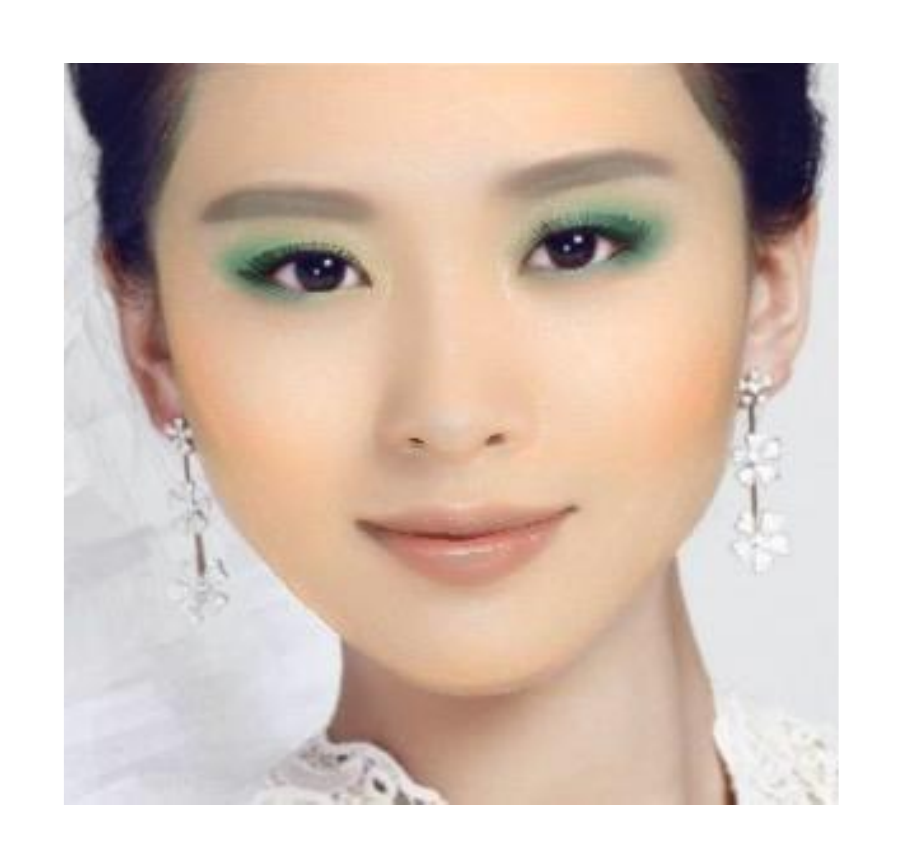}}	
	\caption{The makeup transfer effect of our method. \emph{FIGURE} \ref{fig:Fig1-1}: A input image provided by an ordinary user. \emph{FIGURE} \ref{fig:Fig1-2}: A reference image with reference makeup style. \emph{FIGURE} \ref{fig:Fig1-3}: The output image of our method, where the makeup in \emph{FIGURE} \ref{fig:Fig1-2} was successfully transferred to \emph{FIGURE} \ref{fig:Fig1-1}.}	
	\label{fig:Fig1}
\end{figure*}

Our method based on the technical application of facial makeup transfer completely considered all of the above circumstances. As shown in \emph{FIGURE} \ref{fig:Fig1}, with the image prototype (\emph{FIGURE} \ref{fig:Fig1-1}) as the input image, with the pattern example (\emph{FIGURE} \ref{fig:Fig1-2}) as the reference image, our method could successfully transfer the reference image makeup to the input image to generate output image (\emph{FIGURE} \ref{fig:Fig1-3}).

\section{Related Work}
In 2007, Tong et al. \cite{ref1Tong2007} of the Hong Kong University proposed a facial-to-facial makeup transfer method based on a quotient image. Using the quotient image from a pair of images of the identical person applying and removing makeup as reference images to transfer the reference makeup to the input facial image. Their presented method could be divided into four steps, firstly removing the eyebrows and eyelashes of the input image to prepare for the eye makeup transfer. Then filling resulting holes using texture-synthesis, thus to extract inherent skin features of the input image. Manually specifying the point correspondence between the facial image and the facial model containing 84 landmark points to prepare for facial deformation. Secondly, the reference facial image is deformed according to the input image. Thirdly, the output is multiplied by the input image to achieve the makeup transfer, where the makeup of the same facial before and after is used to indicate the change of the makeup. Finally, eye makeup requires additional processing, which is generally more complicated, and the color is changeable.

In 2009, Guo et al. \cite{ref2Guo2009} of the Singapore National University proposed a simpler method, not for reference image before facial makeup but for an reference image after facial makeup. The method first performs facial alignment between the input facial image and the reference facial image. Since the information is transferred from pixel to pixel, it needs to be fully aligned before transfering, and then layer is decomposed by the Edge Preserving Smooth Filter. The input image and the reference image are resolved into the following three layers: facial structure layer, facial color layer, and  facial detail layer. The information for each layer of the reference image is transferred to the homologous layer of the input image differently: the facial detail layer is direct transferred; the facial color layer is transferred in alpha hybrid mode. The three composite layers are combined to obtain the resulting image.

In 2015, Li et al. \cite{ref3Li2015} of Zhejiang University proposed a facial image makeup editing method based on intrinsic images. The method uses the intrinsic image decomposition method to directly decompose the input facial image into the illumination layer and the reflectance layer, and then edits the makeup information of the facial image in the reflectivity layer, rather than need reference image, and finally decomposes the previous image. The illumination and shadow layers are combined to obtain a makeup editing effect.

In 2016, Liu et al. \cite{ref4Liu2016} of NVIDIA Research designed a new deep convolutional neural network for makeup transfer, which not only could transfer makeup, eye shadow, lip makeup, but also recommend the most suitable input image’s makeup. The network consists of two consecutive steps. The first step is to use the FCN network to parse the facial and resolve different parts, which are distinguished by different colors. The input image and the facial decomposition image of the input image, and the reference image and the facial decomposition image of the reference image are used as input of the makeup transfer network. According to the characteristics of the facial makeup, eye shadow and lip makeup are processed by different loss functions, and the three are integrated. And adding a part of the retained facial image of the input image to get the final result image.

In 2018, Chang et al. \cite{ref5Chang2018} of Princeton University in the United States proposed the PairedCycleGAN network for transfering the facial makeup of the reference image to the input image. The main idea is to train the generation network $\mathcal{G}$ and the authentication network $\mathcal{A}$ to transfer a specific makeup style. Chang et al. \cite{ref5Chang2018} trained three generators separately, focusing the network capacity and resolution on the unique features of each region. For each pair of images before and after makeup, firstly apply a facial analysis algorithm to segment each facial component, such as eyes, eyebrows, lips, nose, and etc. Finally each component is separately calculated and recombined.

\section{Facial Makeup Transfer}
Our method uses the input image $\mathcal{I}$ which applies facial makeup image and the reference image $\mathcal{R}$ which provides the makeup example style as input, and the result is the output image $\mathcal{O}$ which retains the facial structure of $\mathcal{I}$ while applying the makeup style from $\mathcal{R}$. The notation we used is enumerated in \emph{TABLE} \ref{table1}.

\begin{table*}[htbp]
	\centering	
	\caption{The notations are used in this paper.}
	\label{table1}
	\setlength{\tabcolsep}{3pt}  
	\begin{tabular}{|p{0.07\textwidth}|p{0.42\textwidth}|}   
		\hline  
		Notationl &  Meaning \\ 
		\hline
		$\mathcal{I}$ & Input image \\
		\hline
		$\mathcal{R}$ & Reference image (after warping) \\
		\hline
		$\mathcal{O}$ & Output image \\
		\hline
		$\{ .\}_{s}$ & $\{ .\}^{\prime} s$ Facial structure layer \\
		\hline
		$\{ .\}_{d}$ & $\{ .\}^{\prime} d$ Facial detail layer \\
		\hline
		$\{ .\}_{a}$ & $\{ .\}^{\prime} a$ CIELAB  facial color layer a \\
		\hline
		$\{ .\}_{b}$ & $\{ .\}^{\prime} b$ CIELAB  facial color layer b \\
		\hline
		$\alpha$ & Weight controlling the degree of blending $\mathcal{R}_{a, b}$ and $\mathcal{I}_{a, b}$ in $\mathcal{O}_{a, b}$ \\
		\hline
		$\beta$ & Weight controlling the illumination transfer $\mathcal{R}_{s}$ and $\mathcal{I}_{s}$ in $\mathcal{O}_{s}$ \\
		\hline
		$\mathrm{p}$ & Image pixel point \\
		\hline
		$\mathcal{C}_{1}$ & Skin region of the facial image \\
		\hline
	\end{tabular}
\end{table*}

\begin{figure*}[htbp]
	\centering
	\includegraphics[width=7in]{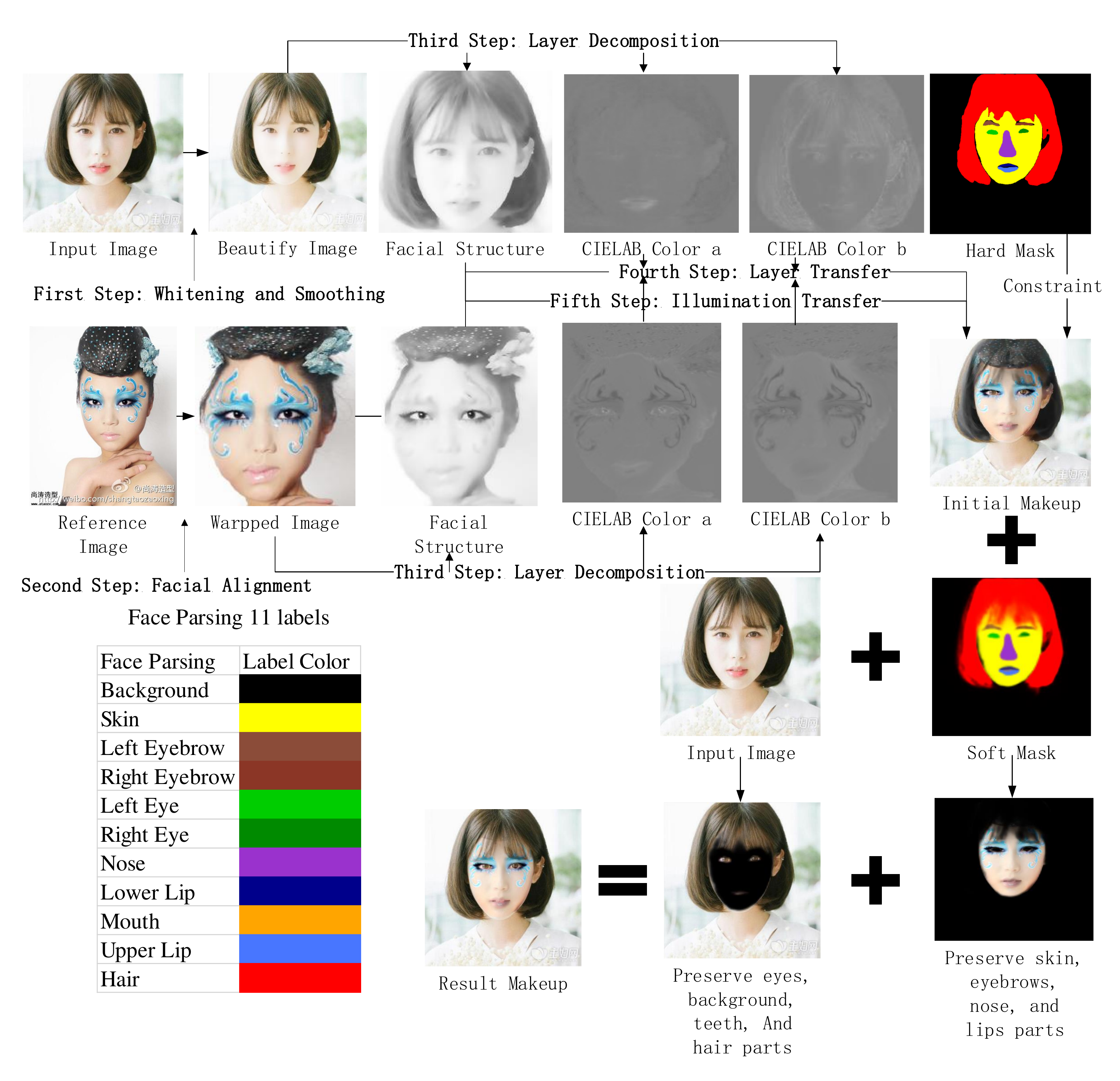}
	\caption{The pipeline of facial makeup transfer. Our method is divided into five steps: whitening and smoothing, facial alignment, layer decomposition, layer transfer and illumination transfer.}	
	\label{fig2}
\end{figure*}

The complete pipeline is shown in \emph{FIGURE} \ref{fig2}. Before the pipeline begins, we need to perform whitening and smoothing pretreatment onto the input image as a small optimization.The pipeline mainly has the following four steps. Firstly, facial alignment has to be done between the input facial image and the reference facial image. Since the information is transferred from pixel to pixel, it needs to be perfectly aligned before the makeup transfer. We use a modified Active Structure Search Algorithm to find the corresponding 90 feature points and affine transformation to distort the reference image $\mathcal{R}$ into the input image $\mathcal{I}$.Secondly, followed by layer decomposition. Both $\mathcal{I}$ and $\mathcal{R}$ are resolved into the following three layers: facial structure layer, facial color layer, and  facial detail layer. Thirdly, the information from per layer of $\mathcal{R}$ is transferred to the related layer of $\mathcal{I}$ in their own way: facial detail is transferred directly; facial color is transferred through alpha blending; facial illumination of the facial structure layer is transferred with specific algorithm. And three composite layers are ultimate combined. Fourthly, we use facial parsing to judge facial label probability of each pixel and then retain the components of the input image and the components of the initial makeup in different probability to fuse into the final makeup.

\subsection{Whitening and Smoothing}
On the one hand, we use the OpenCV Color Balance Algorithm to achieve facial whitening. Color balance global adjustments image dominant colors including red, green and blue. The whole process is briefly described below: firstly initializing image each pixel brightness area (i.e. highlights, mid-tones, shadows), nextly adjusting each brightness area corresponding variable parameters with color balance coefficient, then figuring out image red, green, blue channel value used for adjusting image color, finally balancing the whole image color based on red, green, blue channel value. On the other hand, we use the OpenCV Bilateral Filtering Algorithm \cite{ref12Carlo1998} to achieve facial smoothing. Bilateral filtering performed in the CIELAB color space is the most natural type of filtering for color images: only perceptually similar colors are averaged together, and only perceptually important edges are preserved while eliminating noise. The basic idea underlying bilateral filtering is not only considers the influence of the position on the central pixel, but also considers the similarity degree between the pixel and the central pixel in the convolution kernel, and generates two different weights according to the similarity degree between the position influence and the pixel value. Consider the two weights when computing center pixels, and realize bilateral low-pass filtering.

\subsection{Facial Alignment}
\begin{figure*}[htbp]
	\centering
	\subfloat[Input Image $\mathcal{I}$]{\includegraphics[width=0.3\linewidth,height=0.3\linewidth]{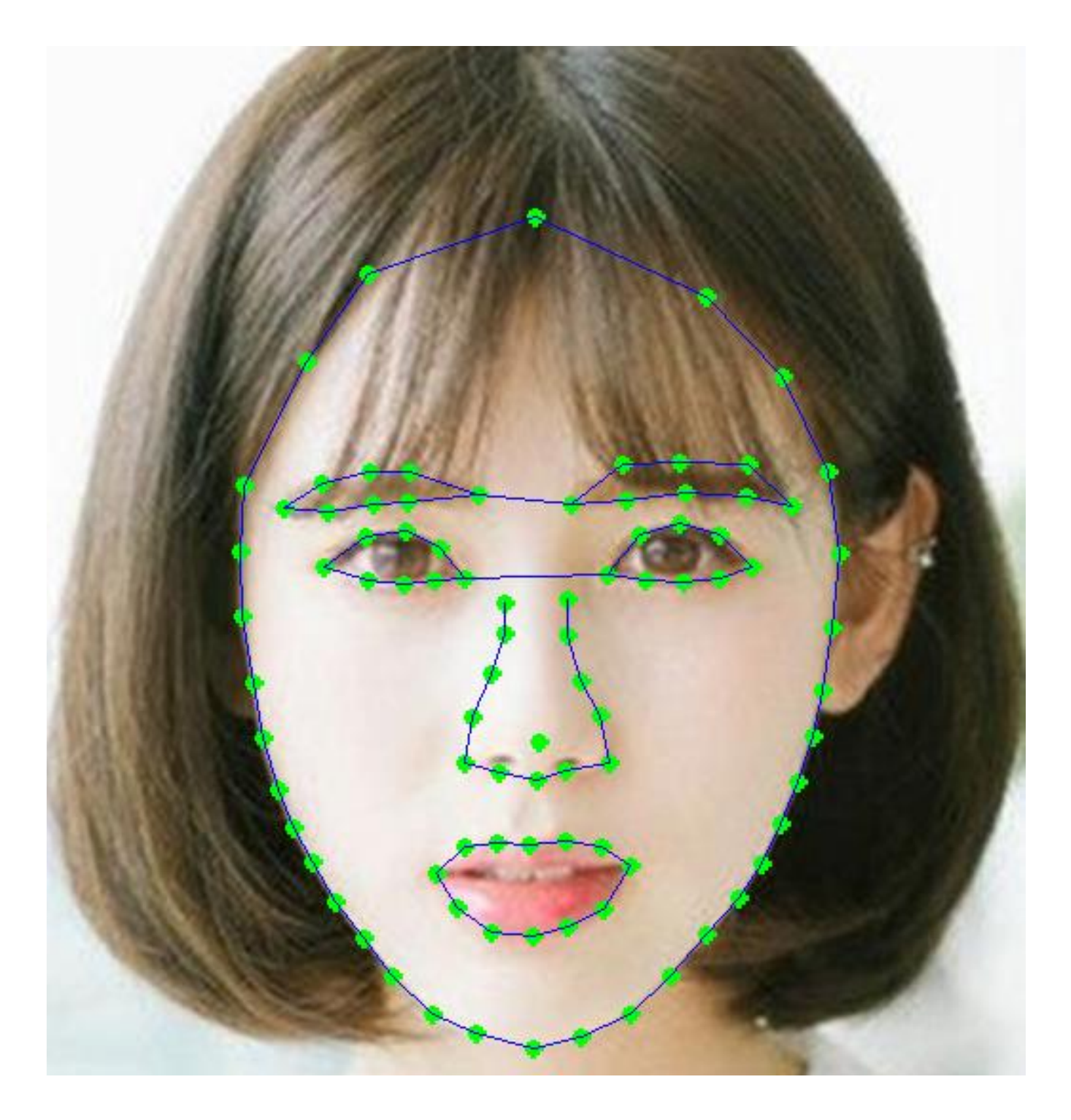}}	
	\subfloat[Reference Image $\mathcal{R}$]{\includegraphics[width=0.21\linewidth,height=0.3\linewidth]{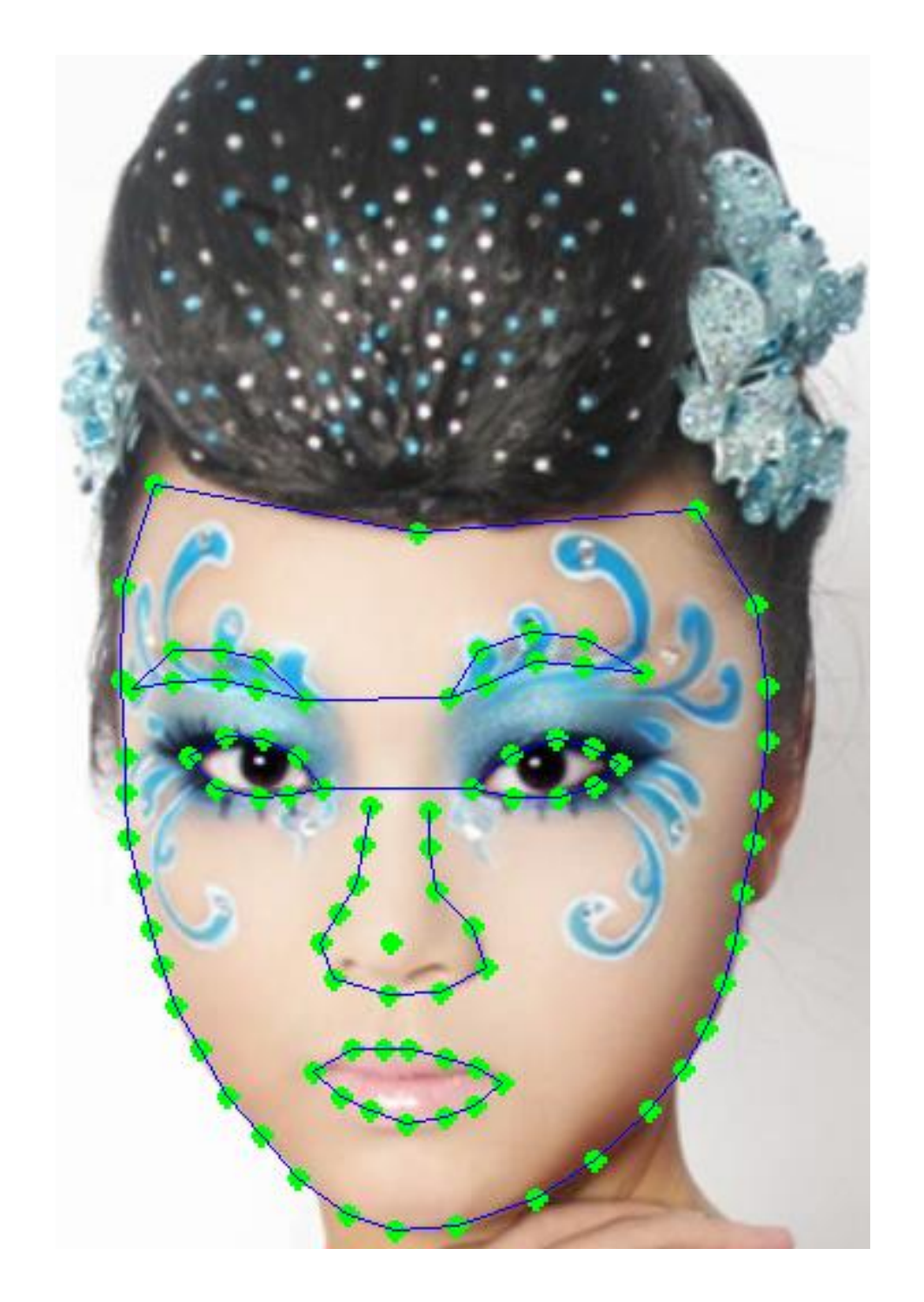}}	
	\caption{Facial feature points landmarked with Active Shape Model (ASM).}
	\label{fig3}
\end{figure*}

For facial alignment, we firstly use the modified Active Shape Model (ASM) of Milborrow et al. \cite{ref14Milborrow2008} to obtain the facial feature points and then use the affine transformation algorithm to warp the reference image $\mathcal{R}$ into the input image $\mathcal{I}$. Due to the variety of appearances in the underside of various possible makeup, our facial feature points landmark software needs to obtain more precise facial feature points in an automatic and manual manner. Our examples of a total of 90 landmark points on the facial are shown in \emph{FIGURE} \ref{fig3}.

\subsection{Layer Decomposition}
\begin{figure*}[htbp]
	\centering
	\subfloat[Input Image $\mathcal{I}$]{\includegraphics[width=0.24\linewidth,height=0.2\linewidth]{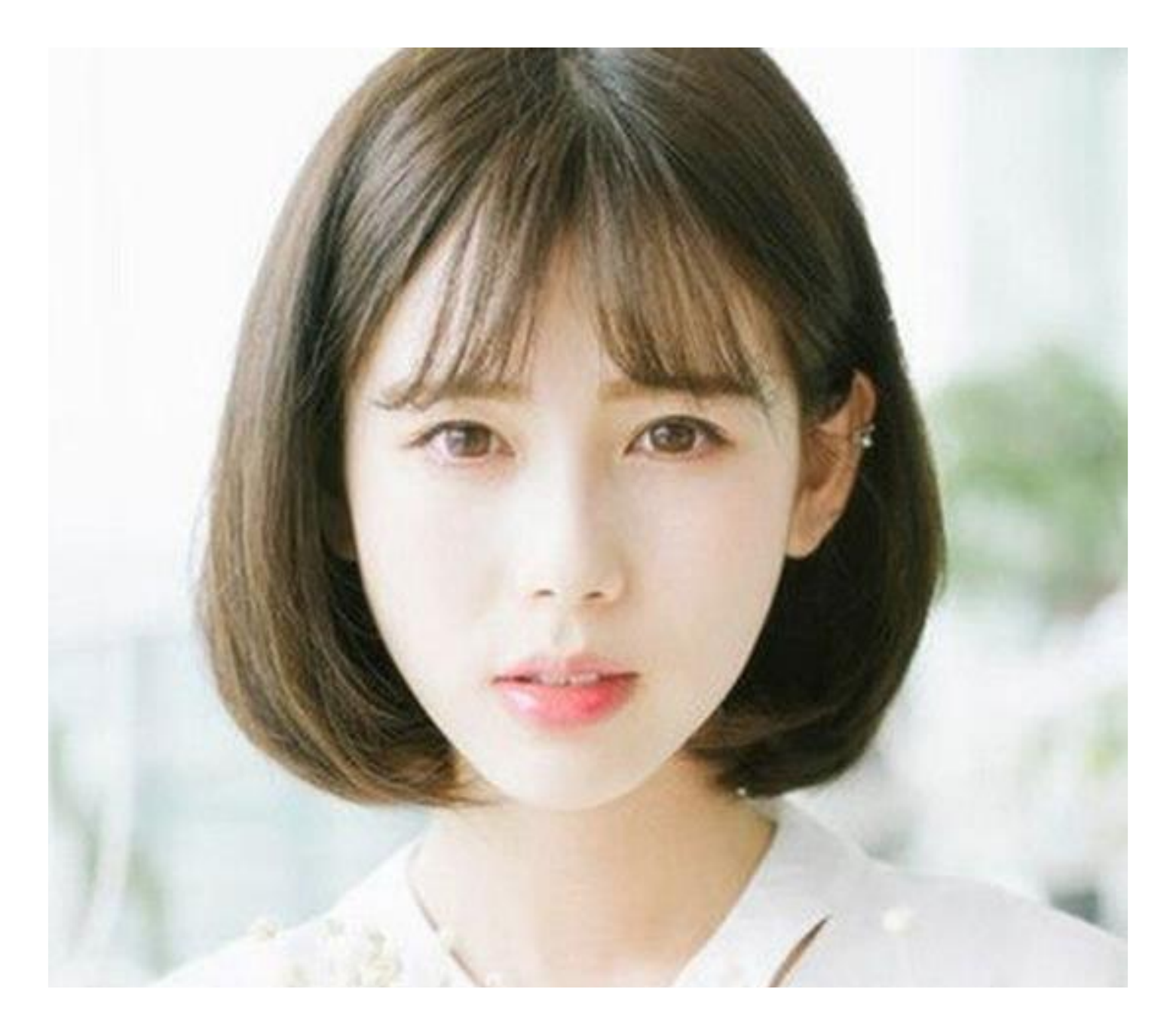}}	
	\subfloat[Facial Label $\mathcal{I}$]{\includegraphics[width=0.24\linewidth,height=0.2\linewidth]{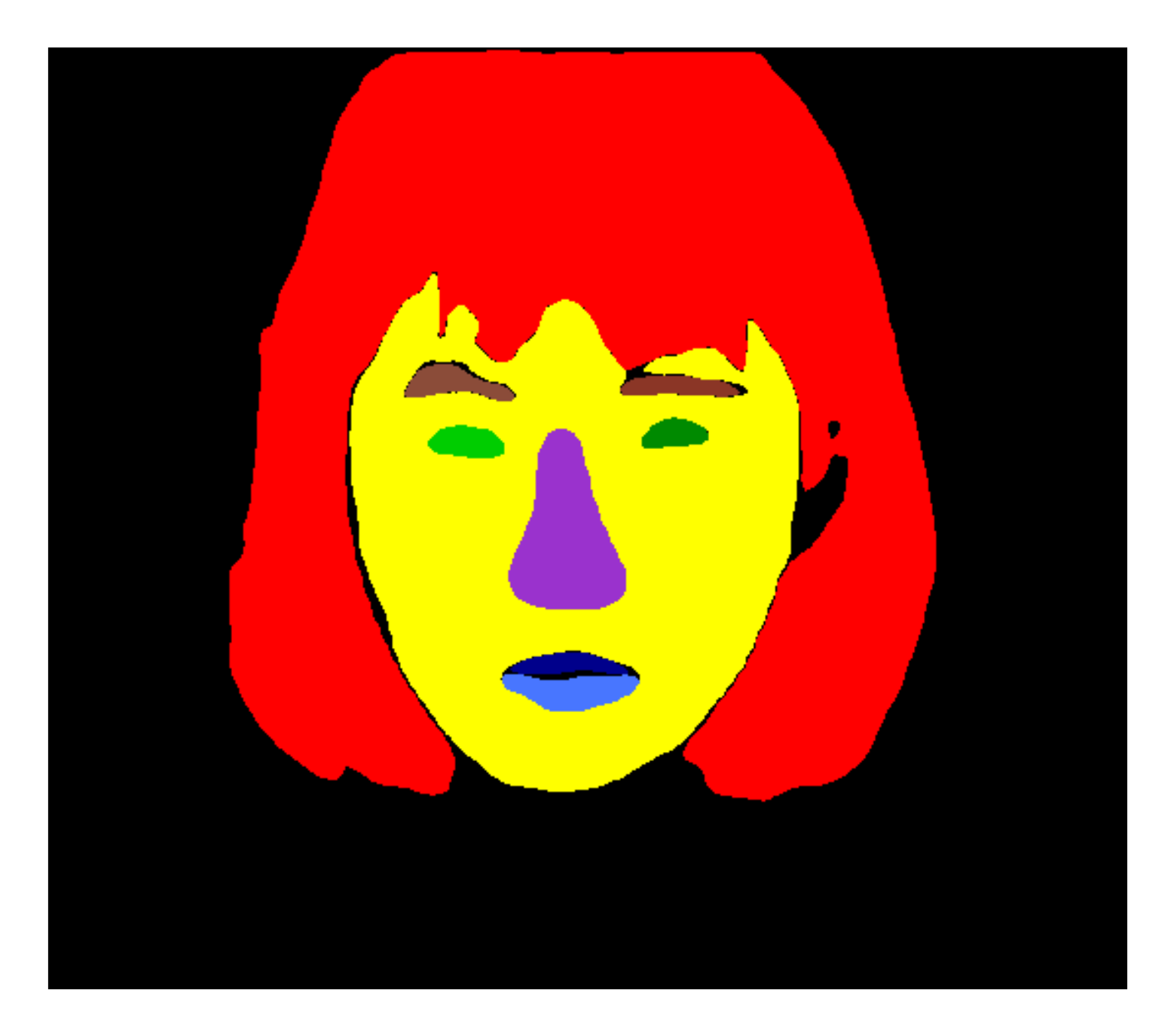}}	
	\subfloat[Reference Image $\mathcal{R}$]{\includegraphics[width=0.2\linewidth,height=0.2\linewidth]{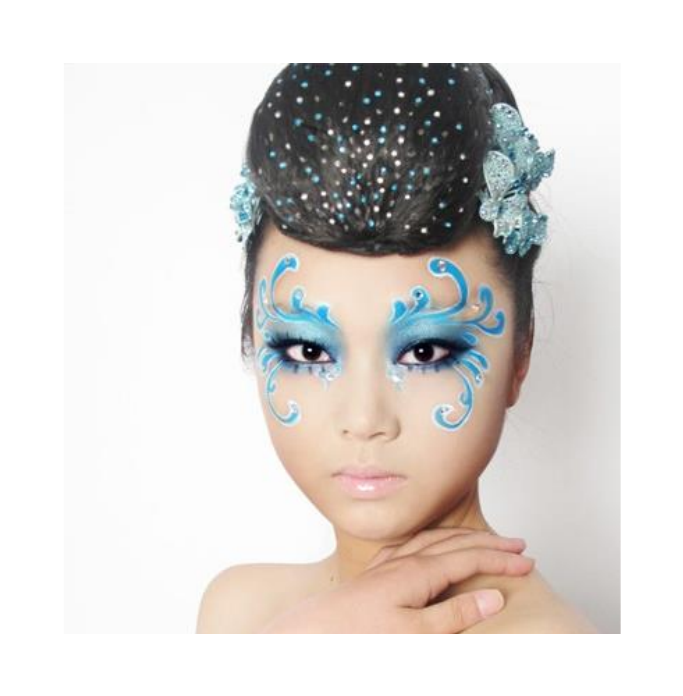}}	
	\subfloat[Facial Label $\mathcal{R}$]{\includegraphics[width=0.2\linewidth,height=0.2\linewidth]{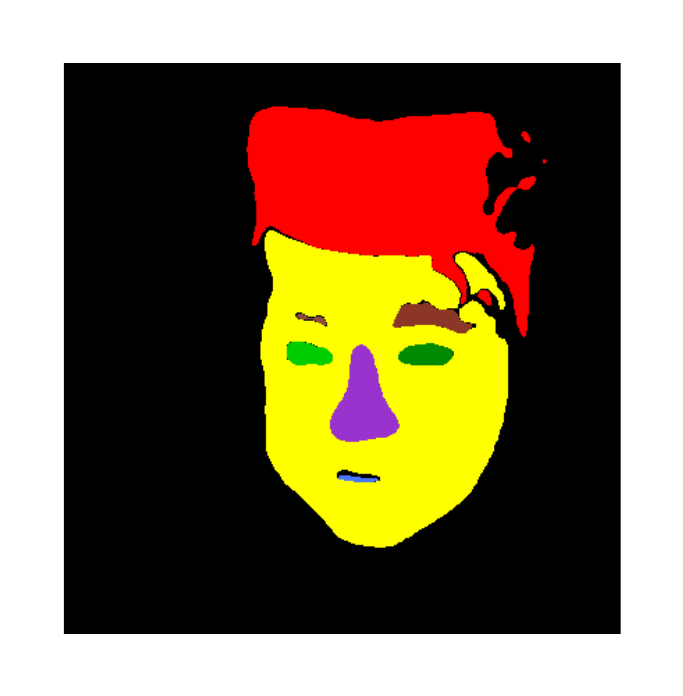}}	
	\caption{Facial components defined by facial parsing of Liu et al. \cite{ref6Liu2015}, including hair, eyebrows, eyes, nose, lips, mouth, facial skin and background.}
	\label{fig5}
\end{figure*}

The facial is segmented according to the components distribution of each pixel. As shown in \emph{FIGURE} \ref{fig5}, we utilize facial parsing of Liu et al. \cite{ref6Liu2015} to define different facial components to obtain components label of per pixel, including hair, eyebrows, eyes, nose, lips, mouth, facial skin and background.

\begin{figure*}[htbp]
	\centering
	\subfloat[Input Image $\mathcal{I}$]{\includegraphics[width=0.24\linewidth,height=0.2\linewidth]{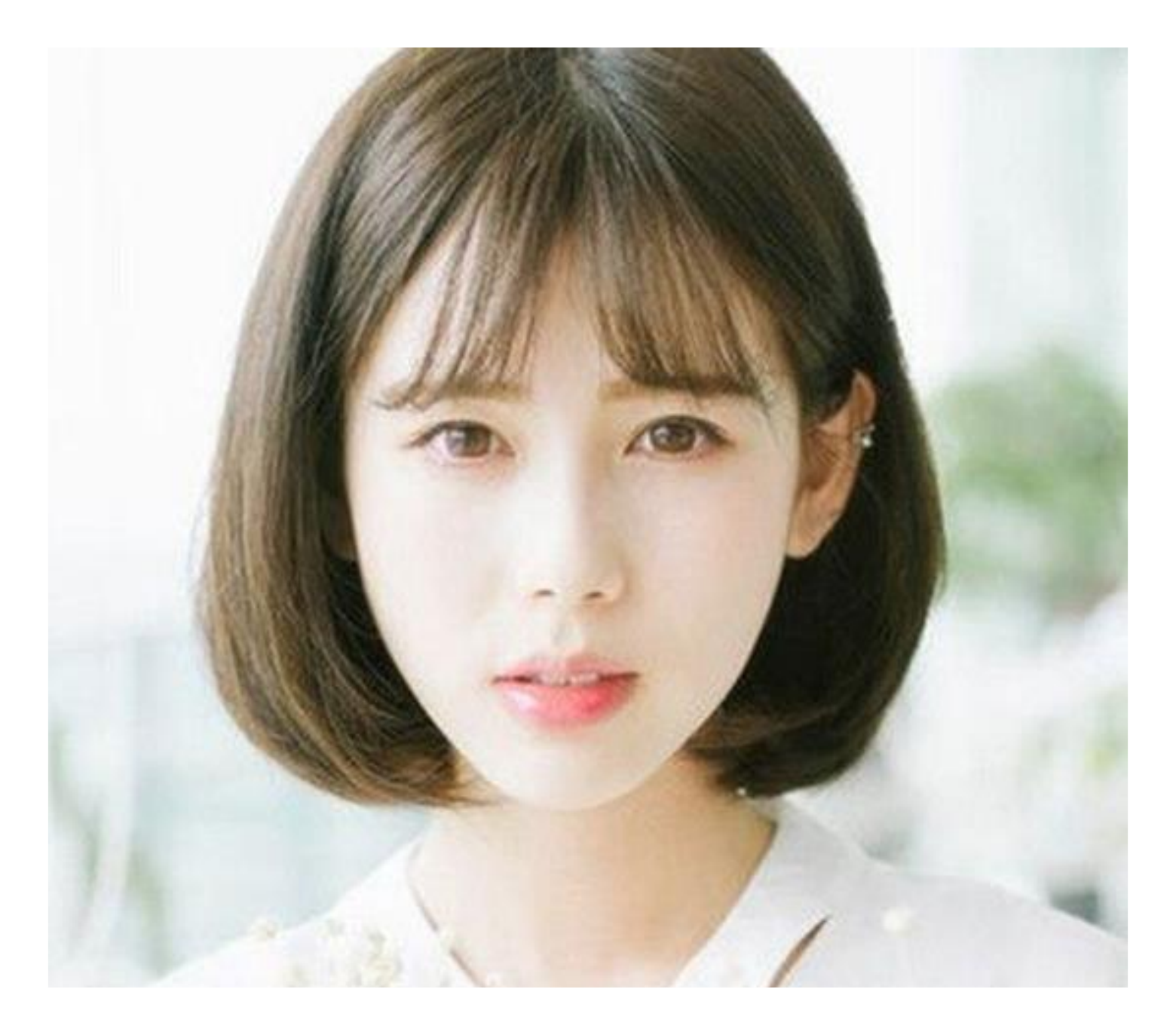}}	
	\subfloat[Landmark Image $\mathcal{I}$]{\includegraphics[width=0.2\linewidth,height=0.2\linewidth]{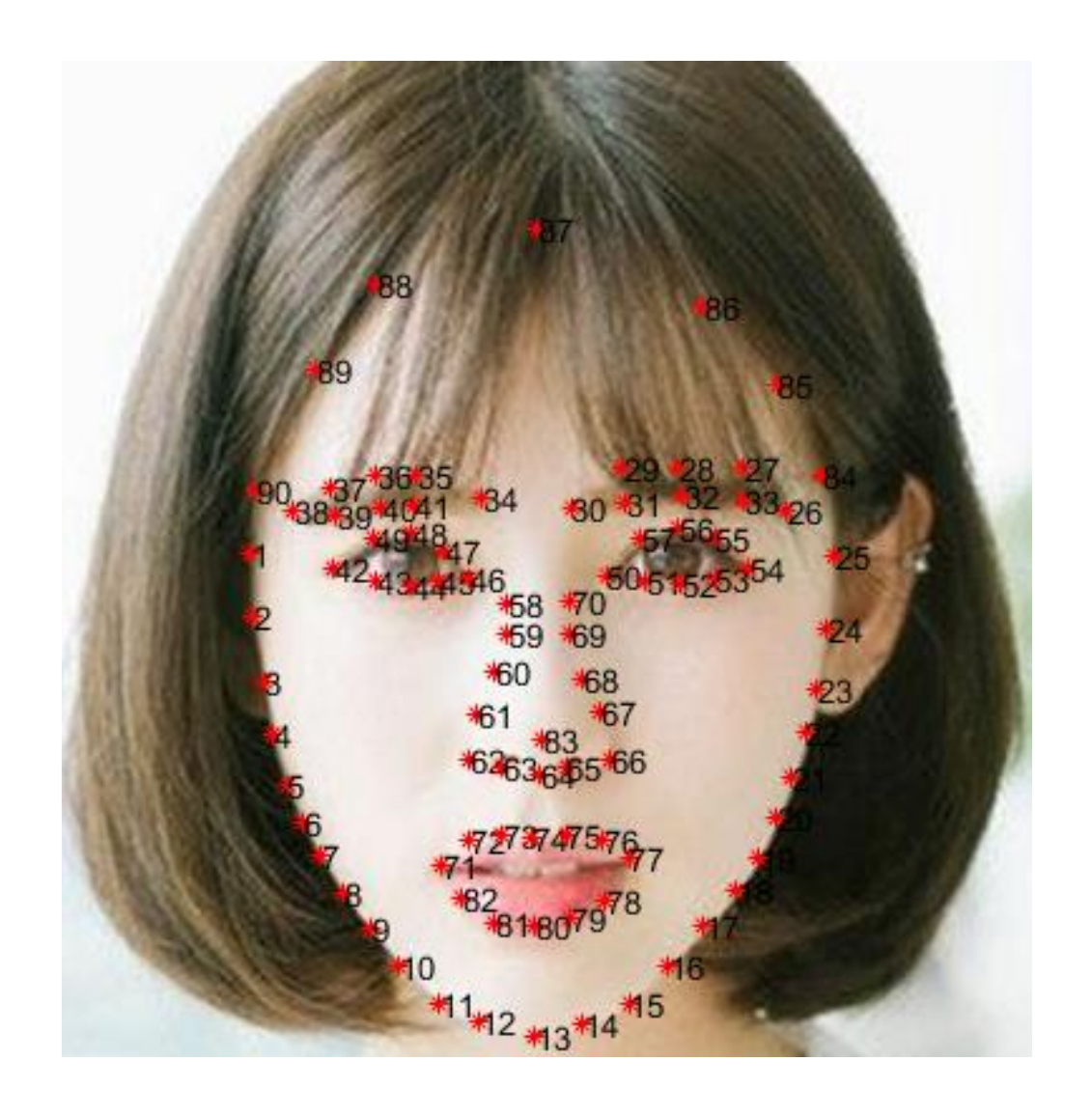}}	
	\subfloat[Reference Image $\mathcal{R}$]{\includegraphics[width=0.2\linewidth,height=0.2\linewidth]{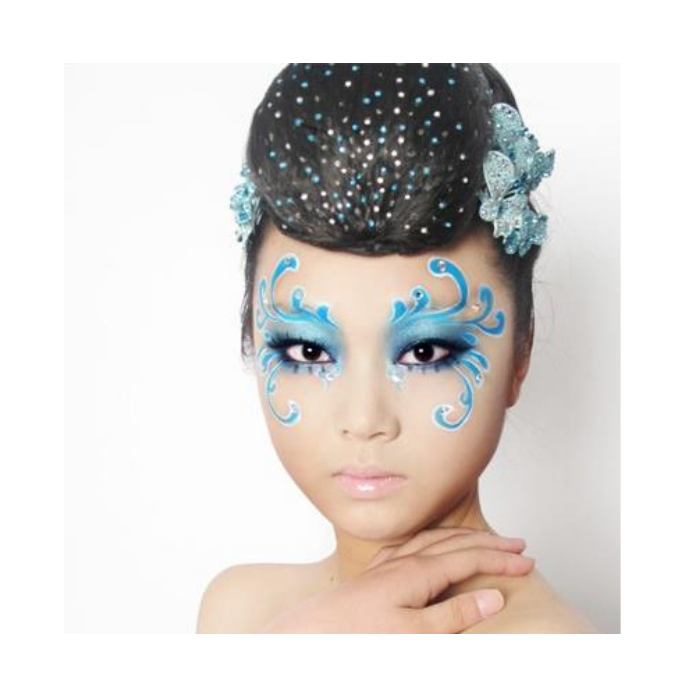}}	
	\subfloat[Landmark Image $\mathcal{R}$]{\includegraphics[width=0.15\linewidth,height=0.2\linewidth]{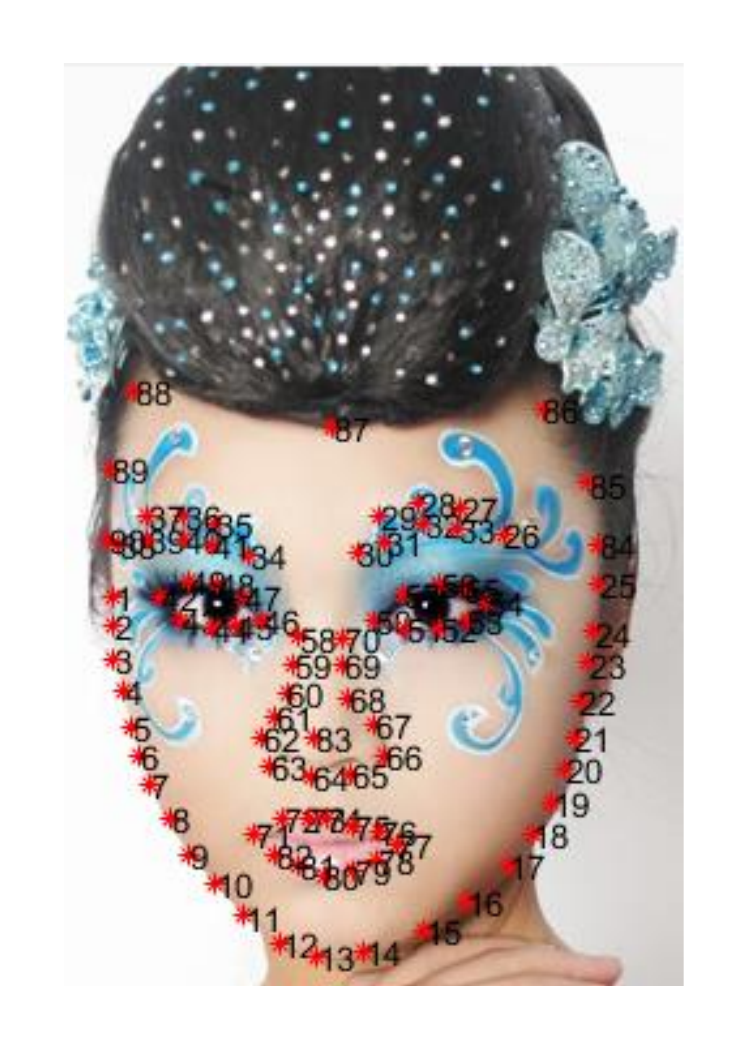}}	
	\subfloat[Warpped Image $\mathcal{R}$]{\includegraphics[width=0.18\linewidth,height=0.2\linewidth]{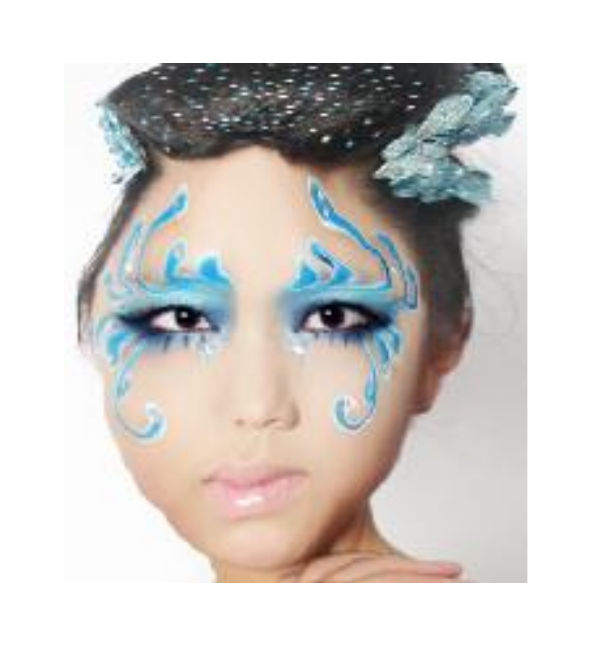}}	
	\caption{Facial alignment by facial warping.}
	\label{fig4}
\end{figure*}

As shown in \emph{FIGURE} \ref{fig4}, we use the above 90 landmark feature points including the input image and reference image to warp the reference image $\mathcal{R}$ to input image $\mathcal{I}$ for facial alignment.

We parse the input facial image and select 11 sorts of labels which seldom cover all the facial components. Then we tint 11 facial component labels to get the facial hard mask. Next we segment facial into different regions with facial hard mask, guiding different makeup transfer operations onto facial regions.

\begin{figure*}[htbp]	
	\centering
	\subfloat[Input Image $\mathcal{I}$]{\includegraphics[width=0.24\linewidth,height=0.2\linewidth]{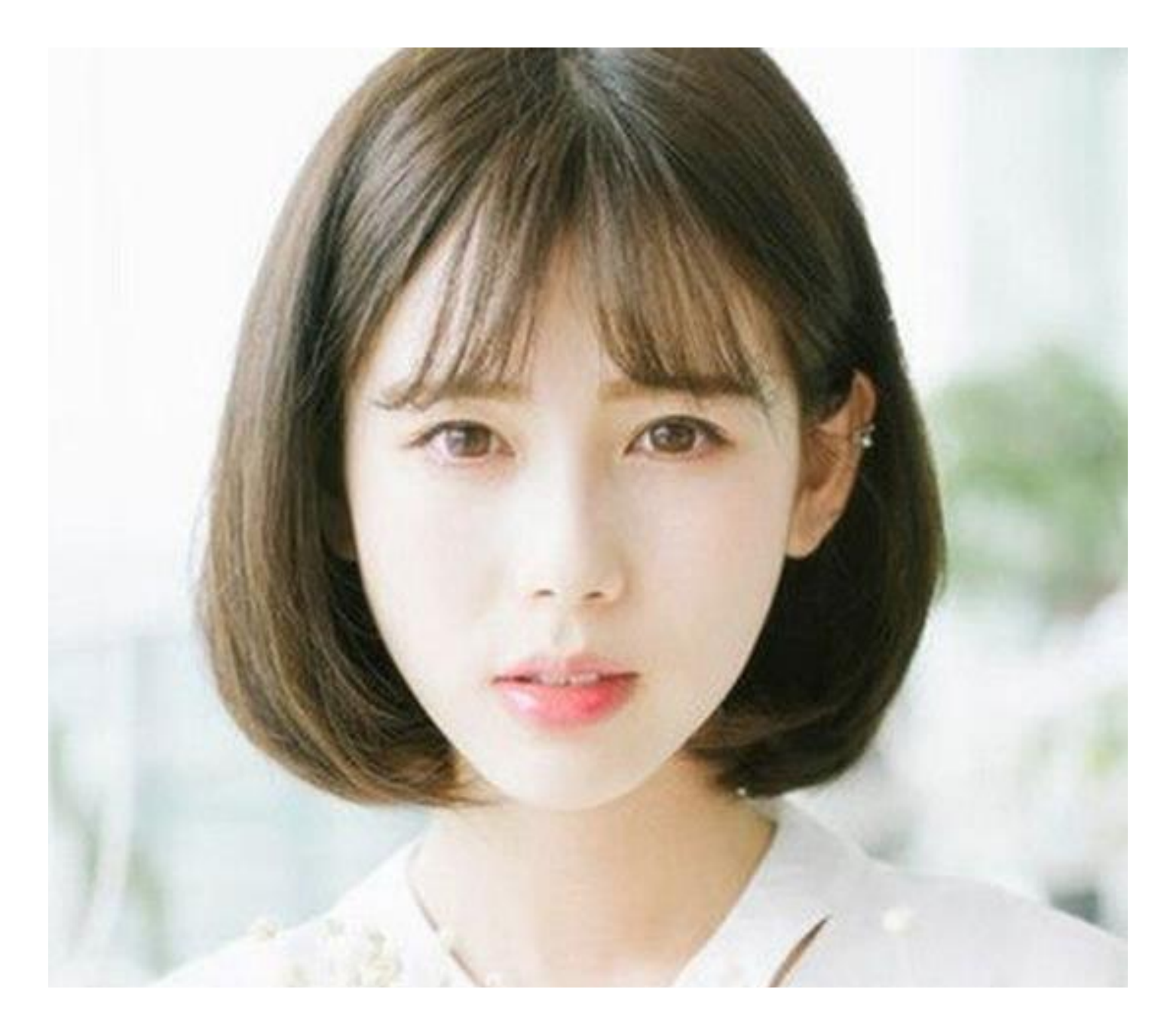}}	
	\subfloat[Facial Mask $\mathcal{I}$]{\includegraphics[width=0.24\linewidth,height=0.2\linewidth]{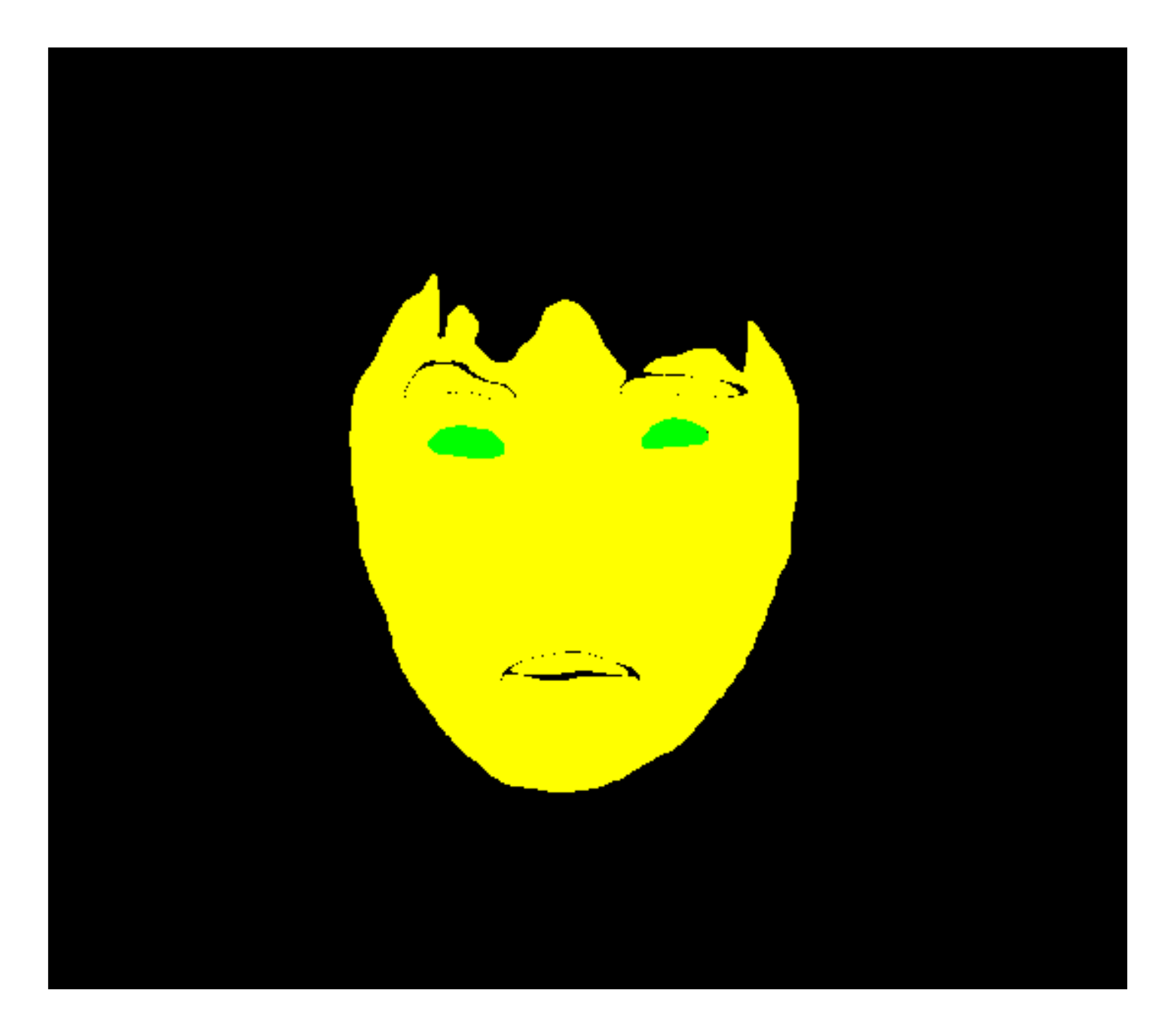}}	
	\subfloat[Reference Image $\mathcal{R}$]{\includegraphics[width=0.2\linewidth,height=0.2\linewidth]{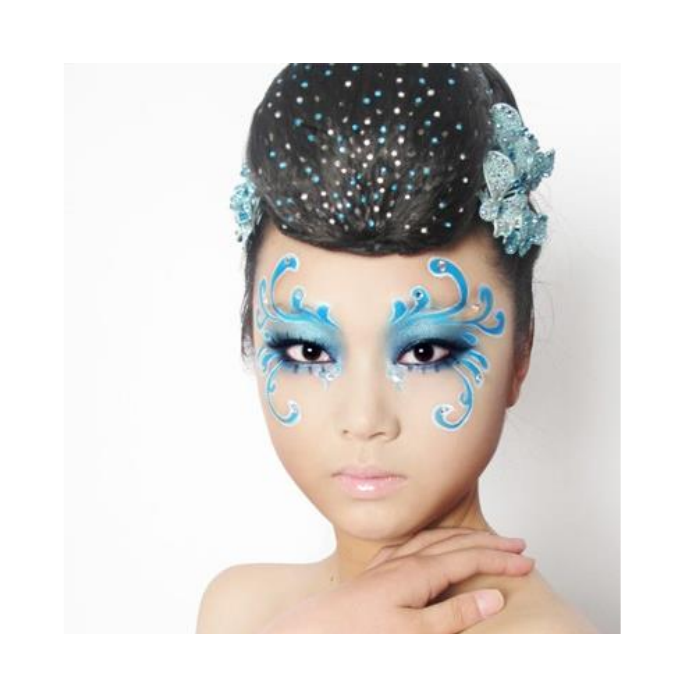}}	
	\subfloat[Facial Mask $\mathcal{R}$]{\includegraphics[width=0.2\linewidth,height=0.2\linewidth]{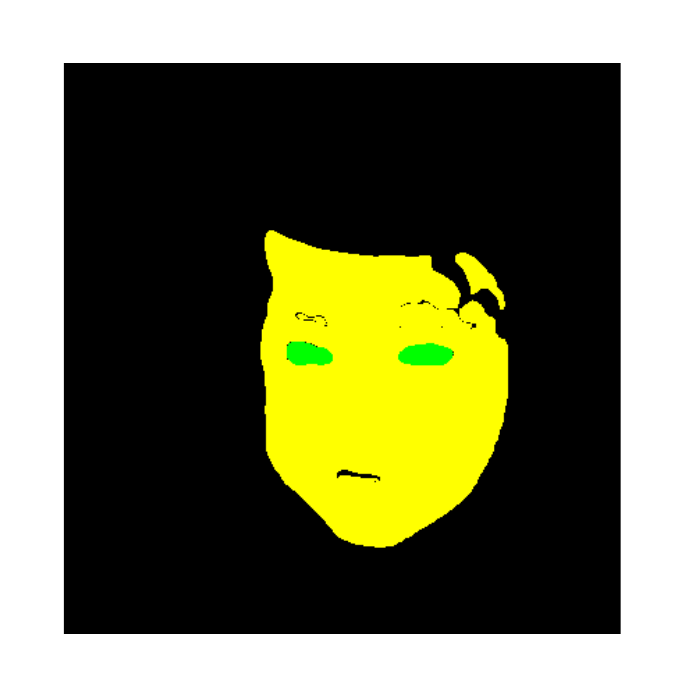}}	
	\caption{To make makeup transfer further achieve better result, we whiten and smooth facial skin component. Then facial components are further divided into three classes, yellow for facial skin as $\mathcal{C}_{1}$; green for eyes and mouth cavity as $\mathcal{C}_{2}$; and the rest of the black part is not considered. Moreover, then the facial part $\mathcal{C}_{1}$ and $\mathcal{C}_{2}$ would perform different makeup task.}
	\label{fig6}
\end{figure*}



We choose CIELAB color space to decompose the input image $\mathcal{I}$ and the reference image $\mathcal{R}$ (after warping) into facial structure layer, facial color layer (i.e. CIELAB color channels a, b channel), and facial detail layer. The CIELAB color space of Lukac et al. \cite{ref7Lukac2007} performs better than other color spaces in terms of separation brightness and approximates the perceptual unity of Wood-land et al. \cite{ref8Woodland2005}.

Secondly, according to the approach of Eisemann et al. \cite{ref9Elmar2004}, Zhang et al. \cite{ref10Zhang2008}, and the Weighted Least Squares (WLS) presented by Farbman et al. \cite{ref1Tong2007}, we perform edge-preserving smoothing filter on the luminance layer $\mathcal{L}$ to extract the facial structure layer ${s}$, then subtracted from the luminosity layer $\mathcal{L}$ to obtain a facial detail layer $d$.

\subsection{Layer Transfer}
We define the facial detail layer $\mathcal{O}_{d}$, i.e.

\begin{equation}
\label{eq3}
\mathcal{O}_{d}=\mathcal{R}_{d}
\end{equation}

We define the facial color layer $\mathcal{O}_{a, b}$ as the alpha-blending of the CIELAB color channels $a$ and $b$ of $\mathcal{I}$ and $\mathcal{R}$, i.e.

\begin{equation}
\label{eq4}
\mathcal{O}_{a, b}(p)=(1-\alpha) \mathcal{I}_{a, b}(p)+\alpha \mathcal{R}_{a, b}(p), \& \mathrm{p} \in \mathcal{C}_{1}
\end{equation}

where $\alpha=0.95$ is the mixing weight that controls the two color channels,$\mathrm{p}$ is the image pixel point, $\mathcal{C}_{1}$ is the skin region of the facial image, and $\& \mathrm{p} \in \mathcal{C}_{1}$ means the image pixel point belonging to facial skin region.

We define the facial structure of $\mathcal{O}_{s}$ as

\begin{equation}
\label{eq5}
\mathcal{O}_{s}=\mathcal{R}_{s}
\end{equation}

\subsection{Illumination Transfer}
We define the following formula to achieve illumination transfer:

\begin{equation}
\label{eq6}
\mathcal{O}_{s}(p)=
\left\{
\begin{array}{c}
\begin{aligned}
	\mathcal{I}_s(p)-\left(\mathcal{I}_{s}(p)-\mathcal{R}_s(p)^{2}\right) / \beta  &, \text { if } \mathcal{I}_s(p)>\mathcal{R}_s(p) \\ 
                                                        \mathcal{I}_s(p) &, \text { otherwise }     \\ 
                                                               &, \& \mathrm{p} \in \mathcal{C}_{1}
\end{aligned}
\end{array}
\right. 
\end{equation}

where $\beta=30$ as the illumination transfer parameter between input facial structure and reference facial structure, $\mathrm{p}$ is the image pixel point, $\mathcal{C}_{1}$ is the skin region of the facial image, and $\& \mathrm{p} \in \mathcal{C}_{1}$ refers to the image pixel point belonging to the facial skin region.

\section{Experiments and Results}

\subsection{Data Collection}
For our makeup transfer experiments, in order to achieve better results, we collect two separate high-resolution datasets, one containing before-makeup faces with nude makeup or very light makeup and another one containing faces with a large variety of facial makeup styles. To this end, we collect our own datasets from major websites. We manually identify whether each facial image is indeed a before-makeup or with-makeup face with eyes open and without occlusions. By this way, we harvest a before-makeup dataset of 526 images and a with-makeup dataset of 878 images. Our datasets contain a wide variety of facial makeup styles.

\subsection{Efficient Makeup Transfer}

\begin{figure*}[htbp]
	\centering	
	\includegraphics[width=7in]{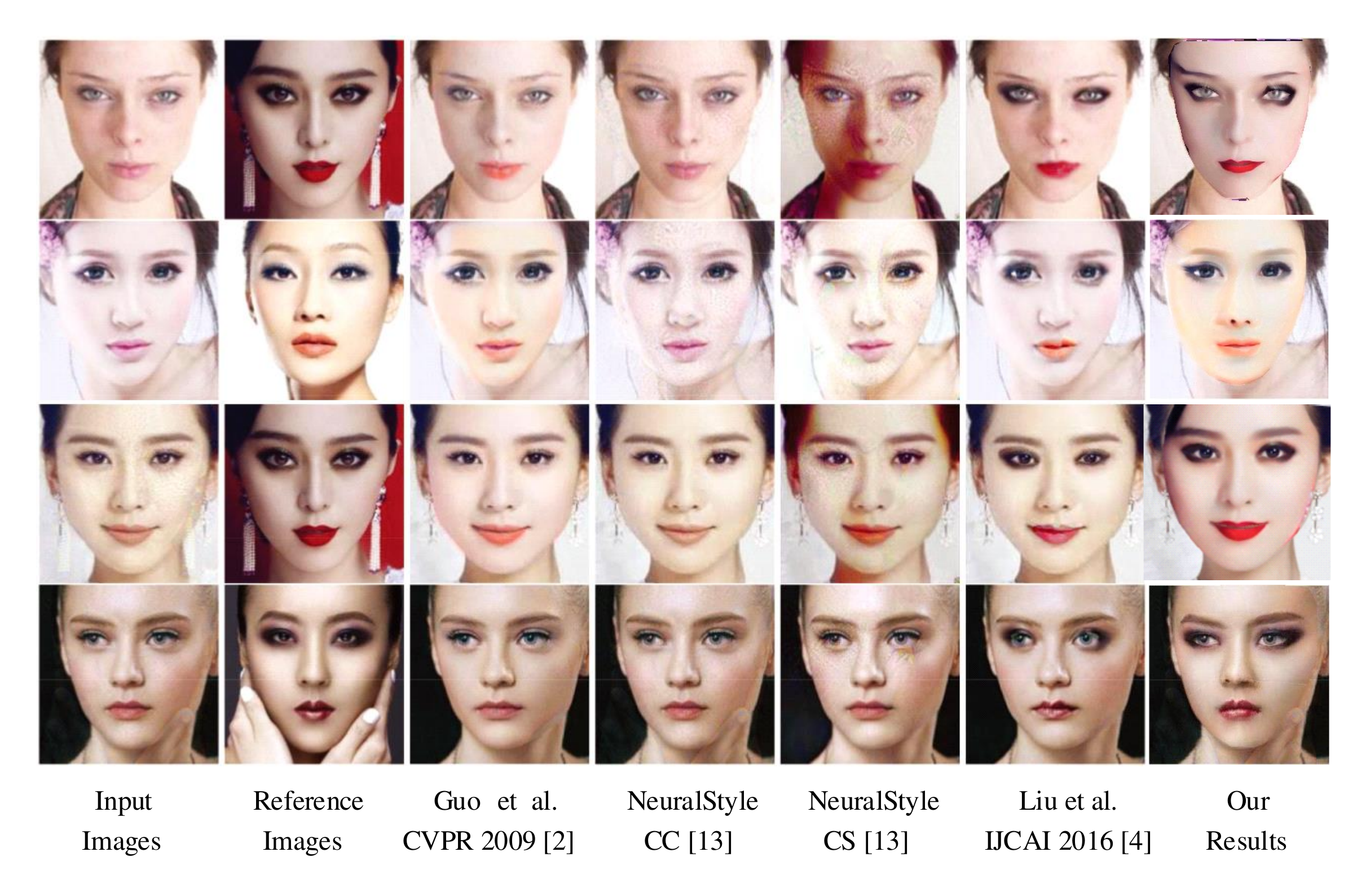}	
	\caption{Comparison results between us and Guo et al. \cite{ref2Guo2009}, Neural Style makeup transfer examples \cite{ref13Gatys2015}, and Liu et al. \cite{ref4Liu2016}. The foundation is uniformly dark, which is not transferred faithfully in Guo et al. \cite{ref2Guo2009} In our result, black or dark makeup appears natural.}
	\label{fig:7}
\end{figure*}


Comparison results between us and Guo et al. \cite{ref2Guo2009} are shown in \emph{FIGURE} \ref{fig:7}. On the one hand, Guo et al. \cite{ref2Guo2009} method assume the illumination in the reference image is uniform, but it is not necessary to be the same as the input image. If any shadow or specularity exists, they would also be transferred to the input image. To solve this problem, we introduce illumination transfer to detect and remove shadow or specularity; our results are shown in \emph{FIGURE} \ref{fig:7}. 

On the other hand, Guo et al. \cite{ref2Guo2009} method does not work well for black and dark makeup. In their result, the dark regions appears gray and unnatural. The black color is the foundation in physical makeup; but their method only transfers the detail introduced by foundation. The black color is interpreted as no color in CIELAB color space; the illumination of black color is especially important to human perception. But the illumination is not transferred in their method. Thus, the dark color in their result appears gray. We solve the problem through the way that adding illumination transfer with user control coefficient in the degree of illumination transfer, and our results are shown in \emph{FIGURE} \ref{fig:7}.

\subsection{Effective Makeup Transfer}

\begin{figure*}[htbp]
	\centering	
	\includegraphics[width=5.3in]{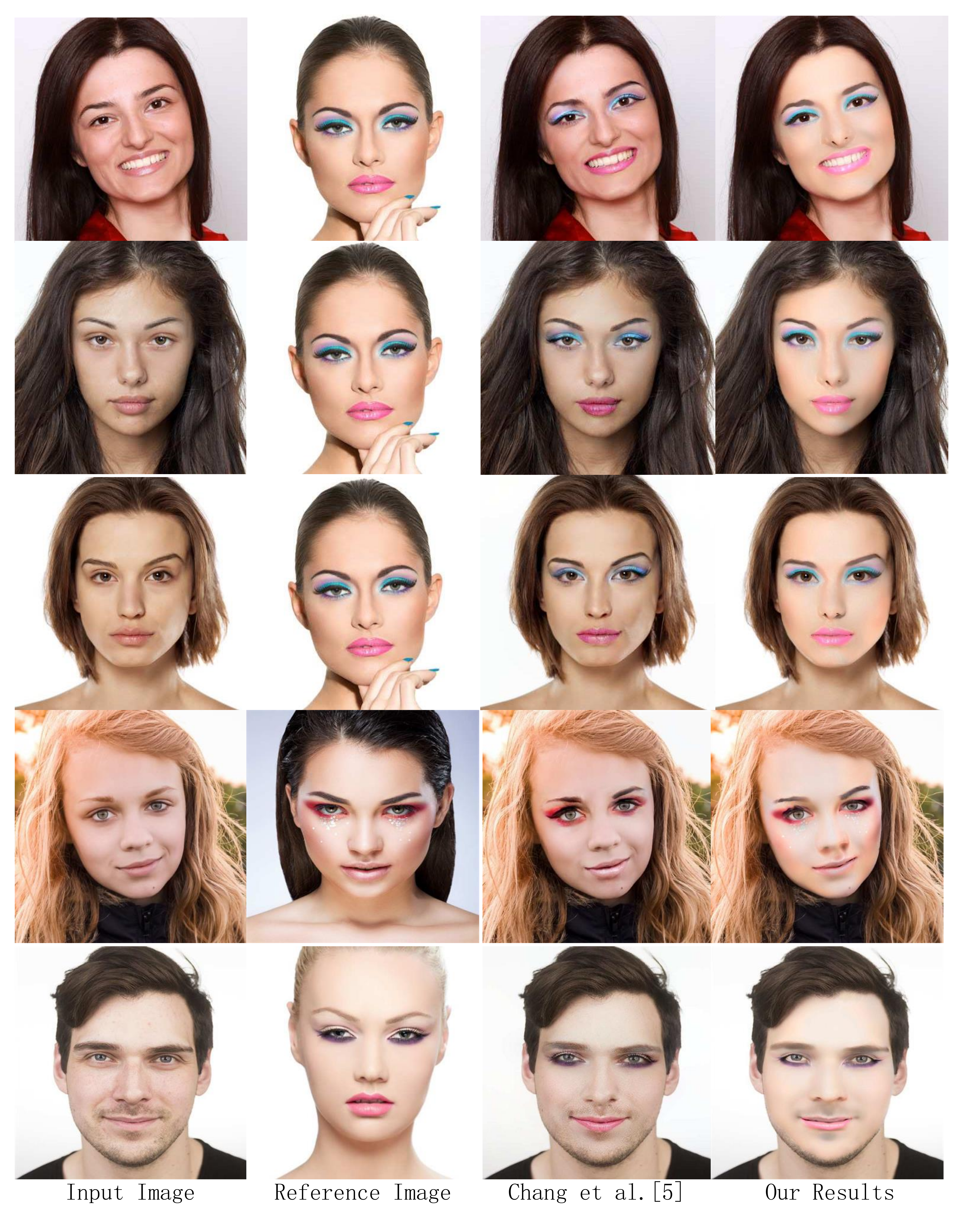}
	\caption{Comparison results between us and Chang et al. \cite{ref5Chang2018}.}
	\label{fig:8}
\end{figure*}


Comparison results between us and Chang et al. \cite{ref5Chang2018}, as shown in \emph{FIGURE} \ref{fig:8}. As we have compared in the above results, the method of the Chang et al. \cite{ref5Chang2018} could only transfer the makeup of the eyes and lips, whereas could not transfer the makeup of the skin part, but our method is not only transfer the eyes and lips makeup, but also transfer the skin part makeup, which is equivalent to a combination of both. Except that, their makeup method for fine hair could not be effectively treated, but our method could overcome it.

Other comparison results between us and Liu et al. \cite{ref4Liu2016}. As shown in \emph{FIGURE} \ref{fig:7}, our makeup result works better than Liu et al. \cite{ref4Liu2016} method. As we could see, our method could transfer facial skin detail of the reference image, thus conduct to form new detail, while Liu et al \cite{ref4Liu2016} method could not do that. Furthermore, our method that combine makeup and relighting could handle the reference image with eye black and dark makeup rather than Liu et al. \cite{ref4Liu2016}.

Last but not least important, the time and space complexity of our method is lower than Liu et al. \cite{ref4Liu2016}.As shown in \emph{TABLE} \ref{table3}, the running time for beautify makeup is within 2 seconds on an iPhone6 for a pair of $224 \times 224$ color image with our method. For Liu et al. IJCAI 2016 \cite{ref4Liu2016}, it needs to take 6 seconds on a TITAN X GPU for a pair of $224 \times 224$ color image.

\begin{table*}[htbp]
	\centering	
	\caption{Runtime Comparison.}
	\label{table3}
	\setlength{\tabcolsep}{3pt}  
	\begin{tabular}{|p{0.16\textwidth}|p{0.28\textwidth}|p{0.03\textwidth}|}  
		\hline  
		Methods  &   Environment &   Time \\ 
		\hline  
		Liu et al. IJCAI 2016 \cite{ref4Liu2016} & $224 \times 224$  image pair using TITAN X GPU & 6s \\
		\hline  
		Our method &  $224 \times 224$  image pair using iPhone6 &  2s  \\
		\hline
	\end{tabular}
\end{table*}

\subsection{Air-Bangs Makeup Transfer}

\begin{figure*}[htbp]
	\centering	
	\includegraphics[width=7in]{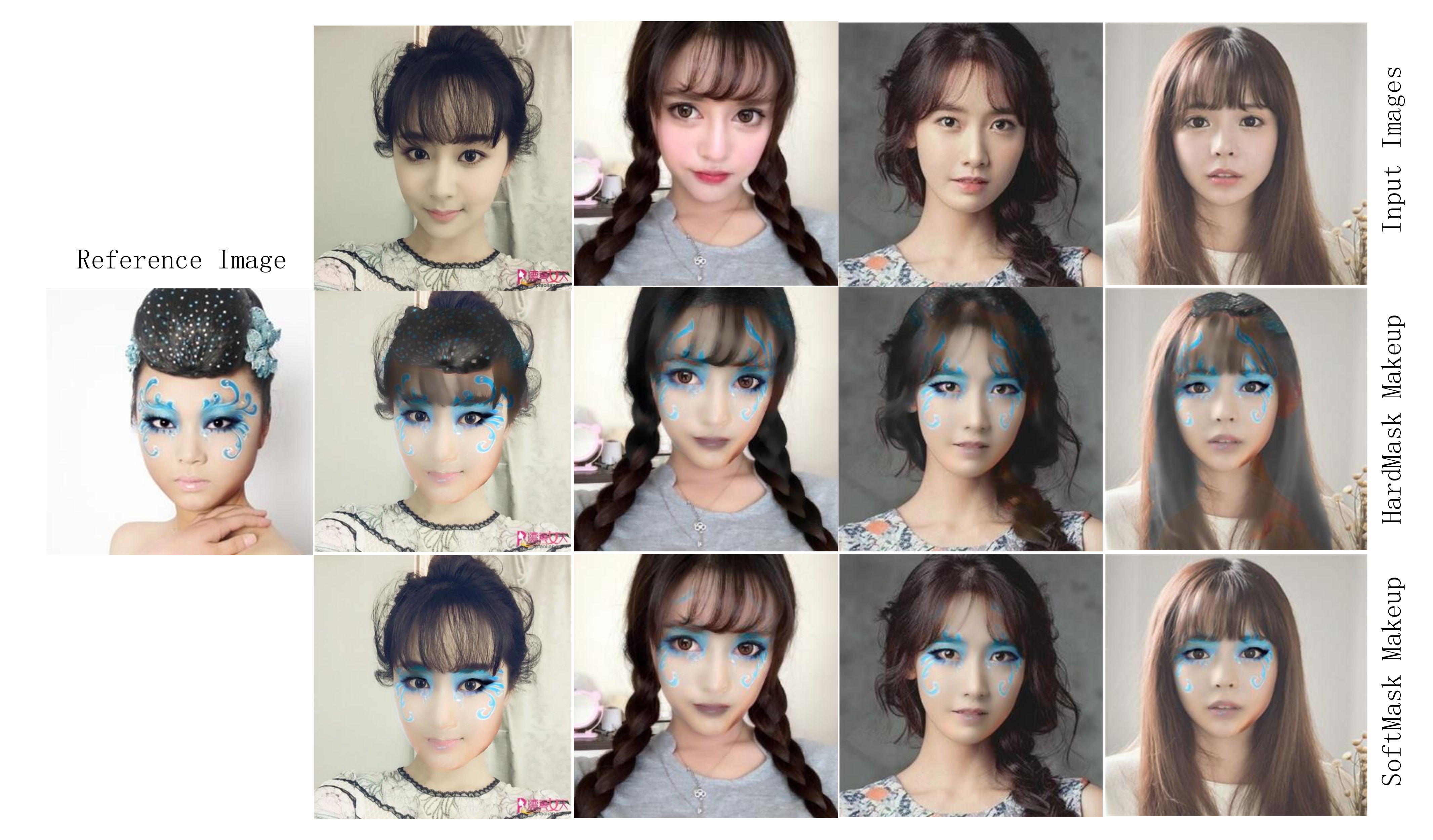}	
	\caption{Examples of our makeup transfer result with air-bangs.}
	\label{fig:9}
\end{figure*}


So far, there is still no good way to deal with makeup transfer with reference examples of air-bangs in the traditional computer vision fields and deep learning fields. Since these methods rely on extremely accurate facial feature landmark without any exception, so as to generate a natural facial mask. As for the reference examples in real life are very diverse, these methods could not make hair and skin very naturally segregate, resulting in the problem that the hair of the reference examples is also transferred together. In order to solve such a tough circumstance, we have further improved our method above, successfully solving the problem of hair and skin boundary in makeup transfer, as shown in \emph{FIGURE} \ref{fig:9}.

The main process as follow, firstly we conduct facial whitening and smoothing and use facial parsing of Liu et al. \cite{ref6Liu2015} to acquire the hard mask of the input image with air-bangs, then we utilize the previous method to generate the initial makeup, in which process we could notice that the hair makeup of the reference image also transfer to the input image unexpected. Followed, we need to convert the hard mask into soft mask which could judge the facial components in terms of probability. Combined the soft mask of the input image, we could make the input image preserve four facial components: eyes, mouth, air-bangs, and the background parts. At the same time, we make the initial makeup preserve four facial components: skin, eyebrows, nose, and lips parts. Thirdly, we fuse the pixels of the input image's facial retention component and the initial makeup result's facial retention component with different probabilities. Finally, we combine the above fusion results to generate the final makeup.

\subsection{Quantitative Comparison}
The quantitative comparison mainly focuses on the quality of makeup transfer and the degree of harmony. On the one hand, we conduct 100 makeup transfer experiments and compare our results with Guo et al. \cite{ref2Guo2009}, Neural Style \cite{ref13Gatys2015}, and Liu et al. \cite{ref4Liu2016}. Each time, a 7-tuple, i.e., a input facial mages, a reference facial image, the result facial images by our method and above methods, are sent to 20 participants to compare. Note that the four result facial images are shown in random order. The participants rate the results into five degrees:“much better”, “better”, “same”, “worse”, and “much worse”. The percentages of each degree are shown in \emph{TABLE} \ref{table4}. Our method is much better than Guo in 23.6{\%} cases. We are much better than NerualStyle-CC and NerualStyle-CS in 90.1{\%} and 92.3{\%} cases. And We are much better than Liu in 32.9{\%} cases. 

\begin{table*}[htbp]
	\centering	
	\caption{ Quantitative comparisons between our method and four other makeup transfer methods. Each percentage in the table means our method is much better (or better, same, worse, much worse) than Guo, NerualStyle-CC, NerualStyle-CS, and Liu in percentage cases.}
	\label{table4}
	\setlength{\tabcolsep}{3pt}  
	\begin{tabular}{|p{0.15\textwidth}|p{0.08\textwidth}|p{0.05\textwidth}|p{0.05\textwidth}|p{0.05\textwidth}|p{0.08\textwidth}|}  
		\hline  
		Methods  &  much better  &  better &  same &  worse &  much worse \\ 
		\hline  
		Guo et al. \cite{ref2Guo2009} & 23.6{\%} & 66.7{\%} & 25.2{\%} & 10.2{\%} & 0.93{\%} \\
		\hline  
		NerualStyle-CC \cite{ref13Gatys2015} & 90.1{\%} & 16.2{\%} & 3.13{\%} & 0.13{\%} & 0{\%} \\
		\hline  
		NerualStyle-CS \cite{ref13Gatys2015} & 92.3{\%} & 16.8{\%} & 1.98{\%} & 0.21{\%} & 0{\%} \\
		\hline  
		Liu et al. \cite{ref4Liu2016} & 32.9{\%} & 28.4{\%} & 5.13{\%} & 0.33{\%} & 0.14{\%} \\
		\hline
	\end{tabular}
\end{table*}

On the other hand, we conduct a user study on Amazon Mechanical Turk making a pairwise comparison among results of the method of Chang et al.[5] and of our method. We randomly select 102 input facial mages and reference facial image, so we have 102 groups of makeup transfer results to compare. Then we ask 10 or more subjects to select which result better matches the makeup style in the reference. On average 87.3{\%} of people prefer our results over those of Chang et al..

\section{Conclusion}
In this paper, we propose a novel makeup transfer method that adapts to most of sample images. The main innovations are as follows: firstly, in the makeup transfer process, we conduct the illumination transfer in the facial structure with our special algorithm; secondly, we expand the makeup to air-bangs circumstances. The major advantages of our method are efficient, effective, and could handle the reference image with air-bangs.

Since the reference images only require skin detail and color information to beautify the appearance, the facial structure of input image is no longer needed, helping to protect the privacy of the makeup actor. We apply the latest and most fashionable makeup examples to our system so that users could apply virtual makeup to their faces in real time according to individual needs, just like a tailor-made personal beauty salon.

As we dilate above, our approach has the following three advantages: 
\begin{enumerate}[(1)]  
	\item Black or dark and white makeup could be effectively transferred by introducing illumination transfer;
	\item Efficiently transfer makeup within seconds compared to those makeup methods based on deep learning framework;
	\item Examples with the air-bangs could makeup transfer perfectly.
\end{enumerate}

\section{Acknowledgements}
We thank all the editors. reviewers and  Prof. Yebin Liu for his advices. This work is partially supported by the National Natural Science Foundation of China (Grant Nos. 61772047, 61772513), the Open Project Program of State Key Laboratory of Virtual Reality Technology and Systems, Beihang University (No. VRLAB2019C03), the Open Funds of CETC Big Data Research Institute Co.,Ltd., (Grant No. W-2018022), the Science and Technology Project of the State Archives Administrator (Grant No. 2015-B-10), and the Fundamental Research Funds for the Central Universities (Grant Nos. 328201803, 328201801). Parts of this paper have previously appeared in our previous work. This is the extended journal version of the conference paper: X. Li, R. Han, N. Ning, X. Zhang and X. Jin. Efficient and Effective Face Makeup Transfer. The 4th International Symposium on Artificial Intelligence and Robotics (ISAIR), Daegu, Korea, 20-24 August, 2019.

\EOD

\end{document}